\documentclass[fleqn,10pt]{wlscirep}
\usepackage[utf8]{inputenc}
\usepackage[T1]{fontenc}
\usepackage{svg}
\usepackage{subcaption}
\usepackage[ruled,vlined]{algorithm2e}

\title{Consistent Point Matching}

\author[1*]{Halid Ziya Yerebakan}
\author[1]{Gerardo Hermosillo Valadez}
\affil[1]{Siemens Medical Solutions, Malvern, USA}
\affil[*]{halid.yerebakan@siemens-healthineers.com}

\begin{abstract}
This study demonstrates that incorporating a consistency heuristic into the point-matching algorithm \cite{yerebakan2023hierarchical} improves robustness in matching anatomical locations across pairs of medical images. We validated our approach on diverse longitudinal internal and public datasets spanning CT and MRI modalities. Notably, it surpasses state-of-the-art results on the Deep Lesion Tracking dataset. Additionally, we show that the method effectively addresses landmark localization. The algorithm operates efficiently on standard CPU hardware and allows configurable trade-offs between speed and robustness. The method enables high-precision navigation between medical images without requiring a machine learning model or training data.

\end{abstract}
\begin{document}

\flushbottom
\maketitle

\thispagestyle{empty}

\section*{Introduction}

With the advancement of digitalization in healthcare, medical imaging data are accumulating at an accelerated pace. Comparisons between previously acquired scans are increasingly valuable but often require redundant navigation to the same anatomical locations in 3D volumes. Registration methods can address this challenge, but they are computationally demanding. Thus, in practice, they are implemented only in limited use cases. As an alternative, less accurate landmark-based registration approaches remain in use, risking the loss of semantic relationships at key points of interest.

Recent literature on machine learning in medical imaging emphasizes voxel-level representation learning for semantic description \cite{bai2023sam++,vizitiu2023multi,cai2021deep}, facilitating identification of corresponding points via coarse-to-fine maximum similarity approaches. Bai et al. \cite{bai2023uae} demonstrated state-of-the-art results on the Deep Lesion Tracker \cite{cai2021deep, yan2018deeplesion} by applying the consistency heuristic during training and incorporating semantic information. Ongoing work seeks better voxel representations for medical images using self-supervised methods \cite{codella2024medimageinsight}.

As an alternative solution to the matching problem, we introduced a “Point Matching” method that avoids full registration and quickly identifies the corresponding locations between multiple medical images \cite{yerebakan2023hierarchical,weikert2023reduction}, without the need for a machine learning model. It uses a multi-resolution sparse sampling descriptor to effectively capture anatomical information, as evidenced by an organ classifier \cite{yerebakan2024real}. Point matching is not a full registration technique and lacks the robustness of traditional methods. In this paper, we enhance its robustness by applying the consistency heuristic of Bai et al.'s UAE \cite{bai2023uae}, incorporating consistency across multiple resolution levels within the similarity search. Consistency-based point matching improves robustness in the four datasets we tested. Unlike the UAE method, our algorithm requires no training, operates in approximately two seconds on standard CPU hardware, and allows easy adjustment of speed-accuracy trade-offs.

%In the medical imaging machine learning literature, there is a growing trend toward learning voxel representations to describe semantic information \cite{bai2023sam++, vizitiu2023multi, cai2021deep}. These voxel representations can be used to identify the most similar points through coarse-to-fine strategies. Bai et al. demonstrated state-of-the-art results on the Deep Lesion Tracker using this methodology combined with the consistency heuristic during training. The consistency heuristic itself is valuable and can also be utilized during testing. We integrated the consistency heuristic at multiple levels of the scan for point matching, achieving a well-balanced trade-off between computational time and precision. Unlike the study presented by Bai et al. \cite{bai2023uae}, our consistent point matching method does not require training. The algorithm runs in 5 seconds on CPU hardware and offers an easy configuration for speed-accuracy trade-offs.

\section*{Methods}

Consistent Point Matching uses components of Point Matching as building blocks. We briefly describe these components in the next subsection and show how the method can be made more robust by incorporating a consistency heuristic.

\subsection*{Point Matching}

Point matching \cite{yerebakan2023hierarchical} is a hierarchical descriptor search method designed to efficiently identify the corresponding anatomical points between pairs of images. Its core efficiency arises from generating descriptors through simple memory lookups and enabling rapid scaling of the descriptor to different resolutions dynamically. The multi-resolution strategy in both the search and descriptor definition allows the method to effectively capture fine anatomical details as well as broader global context.

\begin{figure}[htbp]
    \centering
    \begin{subfigure}[b]{0.45\linewidth}
        \centering
        \includegraphics[width=0.8\linewidth]{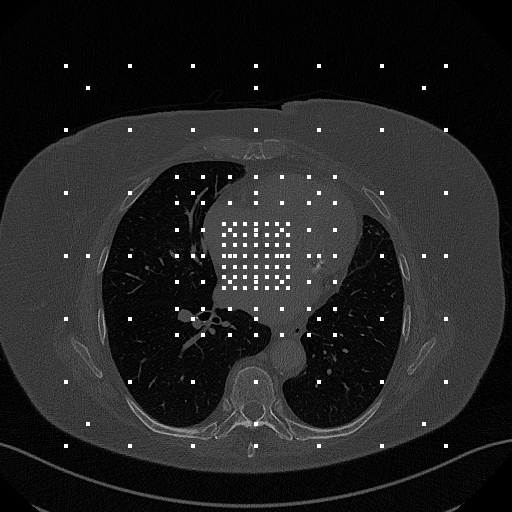}
        \caption{Sparse sampler function on pre-defined offsets}
        \label{fig:sampling}
    \end{subfigure}%
    \hfill
    \begin{subfigure}[b]{0.45\linewidth}
        \centering
        \includegraphics[width=0.8\linewidth]{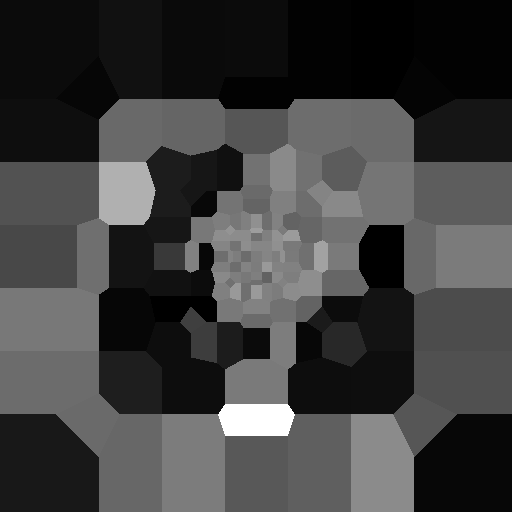}
        \caption{Decoded descriptor of center slice}
        \label{fig:decoded}
    \end{subfigure}
    \caption{Sparse Sampling for Point Matching\cite{yerebakan2023hierarchical} }
    \label{fig:descriptor}
\end{figure}

Descriptors are created by sparse sampling of intensity values based on predefined offset grids, specified in millimeter displacements thanks to reference frame meta data in medical images. Instead of resampling images, the sampler function adapts to the individual voxel spacings, avoiding additional memory and computation. An illustrative example of this sparse sampling and subsequent reconstruction of descriptors is shown in Figure \ref{fig:descriptor} on the center slice of the query. In this specific example, 7x7x7 grids with 8, 20, 48, and 128 mm spacings are used.

%Here, the initial three rows represent orthogonal planes, followed by multiple resolution levels using 9x9x9 grids.

%Point matching\cite{yerebakan2023hierarchical} is a descriptor search strategy to find the corresponding points between pairs of images. The core efficiency comes from curating descriptors with only memory lookups and allowing us to scale descriptor definition on the fly. Thus, the offsets are defined in multiple resolutions that can capture fine details in the query location or global context. Similar anatomical locations produce similar descriptors in terms of cosine and mutual information similarities. 

%The sampler offsets are defined in mm displacements and adapted to images with voxel spacings. In other words, instead of adapting image with resampling it adapts the resampler function to image. Once the offsets are defined, the descriptor is intensity values on the locations where query position is added to offset.  The sampled intensities could be reconstructed back as a 2D image. An exemplary sampling grid is a reconstruction is shown in Figure \ref{fig:descriptor}. In that particular example first 3 rows include 3 orthagonal planes and later ones contains multiple resolution of 9x9x9 grids. 

%Matching points between image pairs are found by maximum similarity. Descriptor similarity between anatomical points is measured using a combination of cosine similarity and mutual information, ensuring robustness across imaging modalities. The method employs a hierarchical search approach that progressively refines the search area across five resolution levels. At each level, a grid of candidate locations is evaluated independently, facilitating parallel computation and achieving significant runtime efficiency without additional hardware acceleration.

The sampler can be easily scaled to perform matching at both fine and coarse resolutions. Offsets defined in millimeters can be multiplied by a scaling factor to enable zoom-in functionality and adjust sampling memory indexes.  The Point Matching method specifically employs five hierarchical levels. At each level, the method iterates over a grid of candidate locations in the target image and selects the best candidate on the basis of a similarity measure that combines mutual information and cosine similarity. Once the optimal point is found, the algorithm advances to the next level, focusing on a smaller, higher-resolution region. Due to the independent calculation of similarities, the method achieves a high degree of parallelization.

\subsection*{Consistent Point Matching}
\begin{figure}[htbp]
    \centering
    \includegraphics[width=0.55\linewidth]{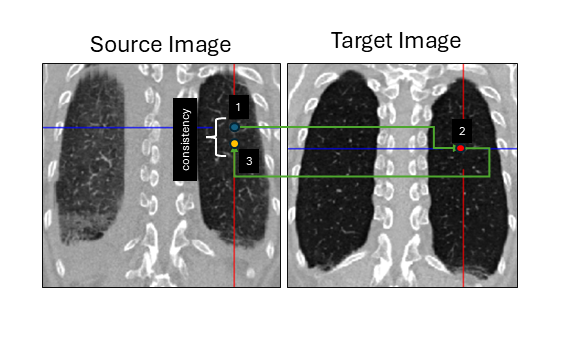}
    \caption{Consistency is distance between original query and round trip estimate}
    \label{fig:consistency}
\end{figure}

Consistency, in our context, is defined as the distance from the original query point to its round-trip estimate back on the source image. First, we map the query to the target image and then map that point back to the source image. If it returns exactly to the original query point, the consistency distance is zero. Figure \ref{fig:consistency} illustrates this process (the yellow dot marks the mapped-back location). Using this metric, the heuristic is that points with lower consistency distances are more likely to be accurate correspondences.
%and estimate location of mapping back the found location in target image to back to query image

We compute the consistencies for multiple nearby locations and incorporate them into the similarity function. Assuming that nearby points have similar offsets in their corresponding positions, we can estimate the required displacement to find the target point. In our implementation, nearby points are selected as six neighbors within a radius of 1.5 and 0.5 times the step size used at the current point matching level (which starts from 16 mm and goes down to 1 mm). Since each neighbor at each scale level votes for the target location, we do not discard estimates; instead, we retain the best five according to the new similarity measure and apply a mean operation to consolidate them. Note that the mapping occurs twice for each of the 12 neighbor points, in addition to the central query, making the naive application 26 times slower. However, with batch processing, the multi-descriptor curation in the search operation is sped up, improving computational time around 2s. The algorithm is given in Listing \ref{algorithm:consistentpointmatching}. In addition to the algorithmic change, we have improved the descriptor definition by adding three orthogonal planes with a resolution of 6 mm using a 7x7 2D grid and an 80 mm 3D grid.

\begin{algorithm}[ht]
\caption{Consistent Point Matching}
\KwIn{query point $Q$, source image $I$, target image $T$}
\KwOut{estimated $\mathrm{center}$ point in $T$}
$s_0 \gets 16$\\
\For{$\ell \gets 1$ \KwTo $5$}{
  $s \gets s_0 \cdot 2^{-\ell}$\;
  $\displaystyle \mathcal{O}\gets \{(0,0,0),(\pm1.5s,0,0),(\pm0.5s,0,0),(0,\pm1.5s,0),(0,\pm0.5s,0),(0,0,\pm1.5s),(0,0,\pm0.5s)\}$\;
  \For{$i\gets1$ \KwTo $|\mathcal{O}|$}{
    $\mathrm{offset}\gets \mathcal{O}[i]$\;
    $F_i \gets \mathrm{PointMatching}(I,\;Q + \mathrm{offset},\;T|center, l)$\;
    $Q'_i \gets \mathrm{PointMatching}(T,\;F_i,\;I|Q, l)$\;
    $d_i \gets \bigl\lVert (Q + \mathrm{offset}) - Q'_i \bigr\rVert$\;
    $\hat{F}_i \gets F_i - \mathrm{offset}$\;
    $w_i \gets \exp\bigl(-d_i/s_0\bigr)\;\cdot\;\mathrm{sim}\bigl(Q+\mathrm{offset},F_i\bigr)$\;
  }
  let $\mathcal{S}$ be the indices of the top-5 $w_i$\;
  $\mathrm{center} \gets \dfrac{\sum_{i\in\mathcal{S}}\,\hat{F}_i}{5}$\;
}
\label{algorithm:consistentpointmatching}
\end{algorithm}

At the first search level, the search range covers the whole image space. Then, the center of the search is updated at each level, and the search space is reduced to a smaller region, similar to point matching.  In the algorithm listing, the function $sim$ stands for similarity, and $F_i$ is the found location in the target image. Each point matching search operation is performed for a single level, since the level loop is taken outside. Alternative formulations for incorporating the consistency distance could be considered; however, simple multiplication with a gaussian kernel $\exp\bigl(-d_i/s_0\bigr)$ is both effective and practical.

\section*{Results}

We evaluated our algorithm on longitudinal matching tasks using CT and MR images. Additionally, we assessed carina landmark location estimation on a CT dataset. We compared consistent point matching with point matching. Images were loaded with positive intensity values and clipped to a range of 0 to 4096. No resampling was applied. %No additional preprocessing is applied images in processing expect positive range representation.

\subsection*{Matching Between Longitudinal Studies}

\begin{figure}[htbp]
    \centering
    \begin{subfigure}[b]{0.45\linewidth}
        \centering
        \includegraphics[width=\linewidth]{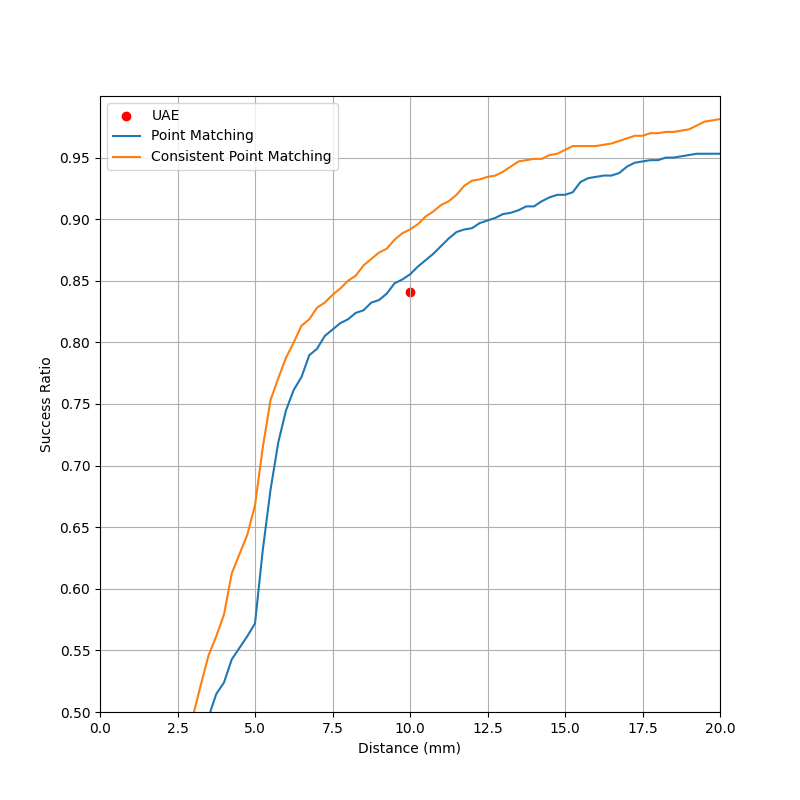}
        \caption{Deep Lesion Tracking Test Annotations}
        \label{fig:dlt}
    \end{subfigure}%
    \hfill
    \begin{subfigure}[b]{0.45\linewidth}
        \centering
        \includegraphics[width=\linewidth]{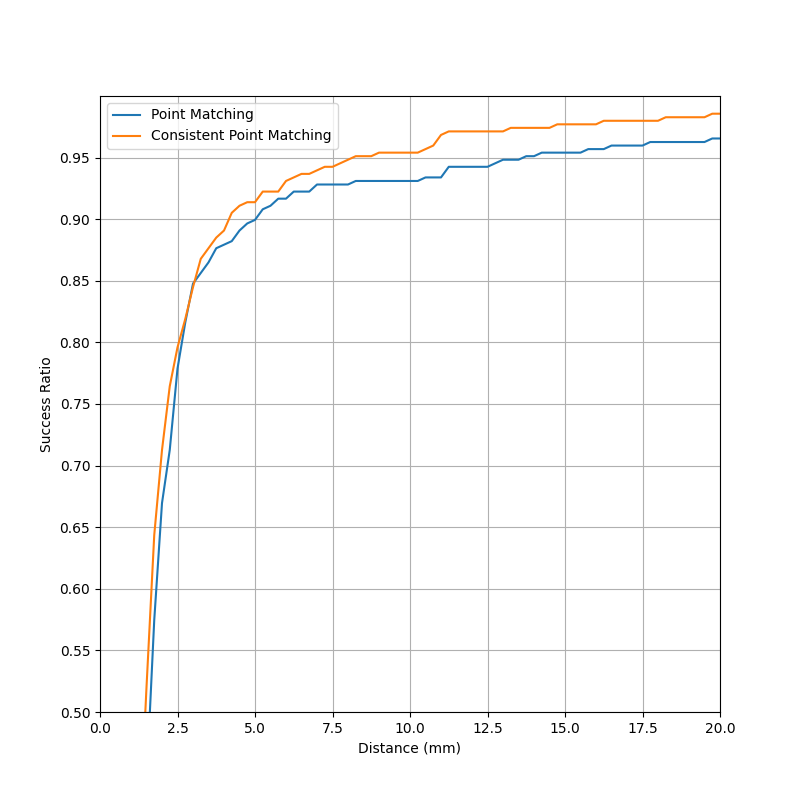}
        \caption{Lung Lesion Annotations}
        \label{fig:nuremberg}
    \end{subfigure}
    \caption{Longitudinal Matching Performance}
\end{figure}

In our first experiment, we used the public DeepLesion dataset with deep lesion tracking annotations provided by \cite{cai2021deep,yan2022sam} that includes different body parts in CT modality. This dataset presents two challenges: first, it has a limited number of slices in the z-axis; second, the slice thickness is 5 mm or greater in 49\% of cases. Only the test set was used, as no training was required. For comparison, we also included the operating point of UAE \cite{bai2023uae}. Although it is a supervised machine learning method on annotated data, it represents prior state-of-the-art performance (0.841@10mm). Since the other methods are compared in this prior work we only compared to UAE.  Despite the challenges, consistent point matching achieves 0.892@10mm, while point matching achieves 0.855@10mm (without radius thresholds). The FROC at different distance thresholds is shown in Figure \ref{fig:dlt}.  The speed per match in this dataset was 1.31s for consistent point matching and 0.16s for point matching, respectively.

%{'PM Precision@10mm': 0.8552083333333333,
% 'CPM Precision@10mm': 0.8916666666666667,
% 'PM Speed': 0.1572935034831365,
% 'CPM Speed': 1.3086965871353944}

In the second experiment, we used an internally curated dataset of 348 location pairs of lung lesions on CT images. Similarly, the FROC curve is obtained by varying the distance threshold to measure the sensitivity of the estimated locations, as shown in Figure \ref{fig:nuremberg}. The algorithm improves the precision and robustness of point matching, achieving 0.954 versus 0.931 at 10 mm. Point matching takes 0.24 seconds per match, while consistent point matching takes 2.26 seconds.

%However, as expected it is slower than original point matching. Each query have 1.5s in avarage vs 0.25s for point matching. 

%{'PM Precision@10mm': 0.9310344827586207,
 %'CPM Precision@10mm': 0.9540229885057471,
 %'PM Speed': 0.2398322772705692,
 %'CPM Speed': 2.266142974639761}

% np.mean(data['Spacing_mm_px_'].apply(lambda x: eval(x)[-1])>=5)

%In our first experiment, we compared the proposed method with respect to
%recently published state-of-the-art results. We utilized Deep Lesion Tracking \cite{cai2021deep,yan2022sam}
%dataset with testing annotations published by [4]. In our case, however, we did
%not utilize the training set since training was not necessary. Also, it is worth to
%note that deep lesion dataset has limited number of slices along z axis which
%makes it more challenging as compared to regular CT scans

%Additionally, we have tested our algorithm on public Deep Lesion Tracking dataset’s test set. The data contains more body regions on lesion type matches. The performance improvement is more pronounced in this dataset where Consistent Point Matching reaches 99 of the matches less than 25mm. 

\newpage
\subsection*{Multi-Modal Study}

We additionally evaluated our method on an in-house study dataset containing multi-time-point CT and MR modalities. This dataset includes aortic aneurysms, intracranial aneurysms (ICA), enlarged lymph nodes, kidney lesions, meningioma, and pulmonary nodule pathologies with a total of 339 pairs of intra-modality matches. In this study, the annotations were provided by multiple annotators. Radiologists were presented with pairs of images along with a description of a predefined finding in the current studies and were asked to find the corresponding locations in previous studies. We used the median of the available annotations as the ground truth.

\begin{figure}[htbp]
    \centering
    \begin{subfigure}[b]{0.45\linewidth}
        \centering
        \includegraphics[width=\linewidth]{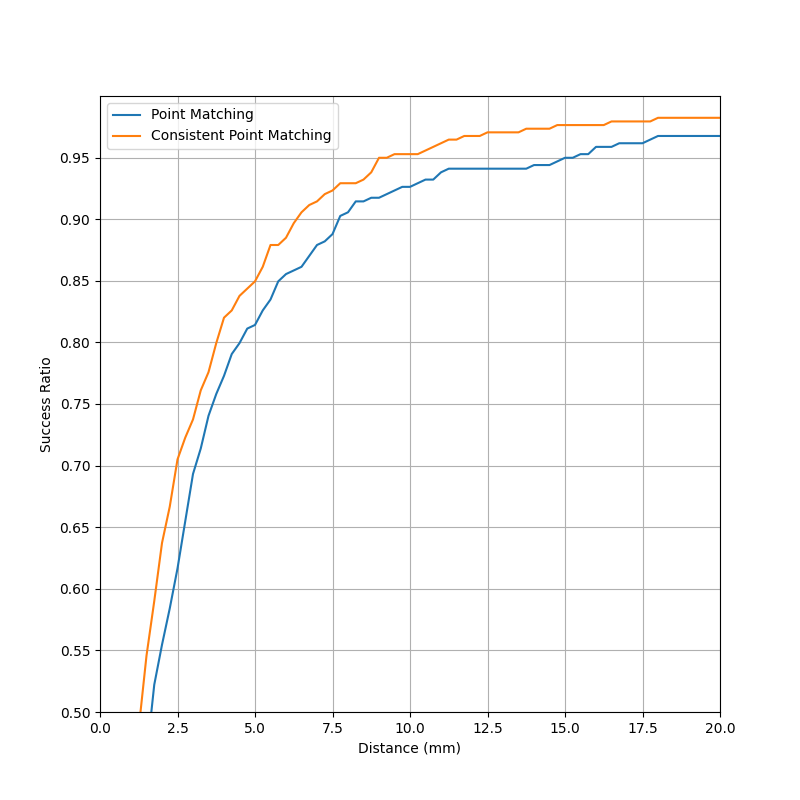}
        \caption{CT, MR followup annotations}
        \label{fig:multimodal}
    \end{subfigure}%
    \hfill
    \begin{subfigure}[b]{0.45\linewidth}
        \centering
        \includegraphics[width=\linewidth]{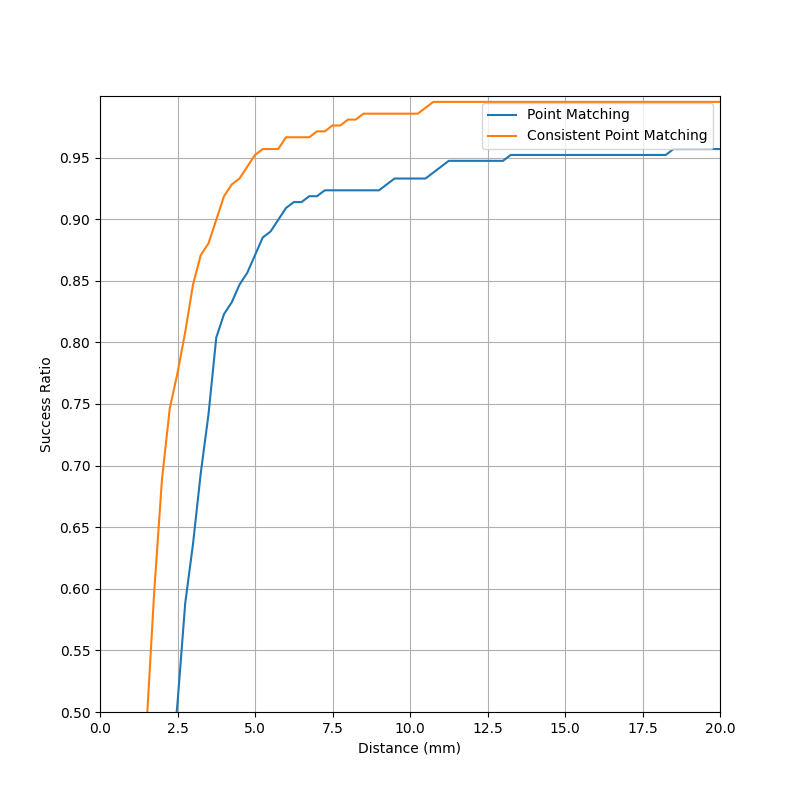}
        \caption{Landmark Annotations}
        \label{fig:landmarks}
    \end{subfigure}
    \caption{Localization performance on different tasks}
\end{figure}

The results indicate that our algorithm generalizes to diseases and modalities. Consistent point matching reaches 95.2 percent at 10 mm, with an average execution time of 2.46 seconds. The FROC curve is shown in Figure \ref{fig:multimodal}.

%{'PM Precision@10mm': 0.9262536873156342,
% 'CPM Precision@10mm': 0.9528023598820059,
% 'PM Speed': 0.2609959907588002,
% 'CPM Speed': 2.460525575288981}

%In our first experiment, we compared the proposed method with respect to
%recently published state-of-the-art results. We utilized Deep Lesion Tracking
%Point Matching 7
%dataset with testing annotations published by [4]. In our case, however, we did
%not utilize the training set since training was not necessary. Also, it is worth to
%note that deep lesion dataset has limited number of slices along z axis which
%makes it more challenging as compared to regular CT scans.

%The measurements are consolidated as ground truth by up to 9 different
%radiologists by taking the median of their annotations. We have used a series
%with more than 3 annotators to compare different time points. Overall, 211 pairs
%of series were selected with this criteria.
%We have estimated the corresponding prior locations of the findings with
%our point matching algorithm compared with expert annotations. We have illus-
%trated the change in sensitivity with different distance thresholds in Figure 2e.
%Each annotator is labeled with a number in this figure. Annotators have better
%localization below the 5mm. However, due to some annotation disagreement in
%a few cases, there are two annotators below the automated algorithm in a larger
%distance range. Notably, our algorithm is very close to an average radiologist
%annotation robustness at 25mm.

\subsection*{Speed Precision Trade off}

In this experiment, we varied the number of consistency points to evaluate the performance-precision trade-off. We used the Deep Lesion Tracking dataset for this purpose as well. Three points include only $stepsize/2$ for the laterality offsets, while seven points include six neighbors. As can be seen in the table, the most robust results are achieved when 13 points are included. However, even with 3 points, there is a significant gain with only a small performance penalty compared to regular point matching, as shown in Table \ref{tab:speed}. The mean distance drops more than the median, indicating an improvement in robustness.

\begin{table}[h!]
\centering
\begin{tabular}{lccc}
\hline
\textbf{Method} & \textbf{Mean Distance (mm)} & \textbf{Median Distance (mm)} & \textbf{Time (s)} \\
\hline
Point Matching         & 5.90  & 3.56  & 0.12 \\
Consistent Point Matching (3)     & 4.82  & 3.16  & 0.41 \\
Consistent Point Matching (7)    & 4.65  & 3.15  & 0.67 \\
Consistent Point Matching (13)        & 4.65  & 3.05  & 1.06 \\
\hline
\end{tabular}
\caption{Mean and median distances, and elapsed time for each method. Lower values are better.}
\label{tab:speed}
\end{table}

%- **Mean Distance**: Lower is better (closer to ground truth).
%- **Mean Speed**: Lower is better (faster).

\subsection*{Landmark localization using Atlas Annotation}
 
Lastly, we used consistent point matching for detecting carina landmarks in CT images. We collected 209 annotations on CT scans for testing and then used a single template landmark on a separate atlas image to find the corresponding locations in other images. We evaluated the localization performance using an FROC curve, as shown in Figure \ref{fig:landmarks}. Quantitatively, point matching achieves 0.933@10mm, whereas consistent point matching achieves 0.985@10mm.

It is certainly easy to scale to multiple templates for increased robustness. However, our experimental results indicate that even with a single template, landmark localization performance approaches the level of supervised landmark detectors. This is due to the consistent structure of the human body.

% {'PM Precision@10mm': 0.9330143540669856,
% 'CPM Precision@10mm': 0.9856459330143541,
% 'count': 209}

%Ablation: Same descriptor, Bigger descriptor

%[Qualitative Results on Landmarks, Estimates and GT]

\section*{Discussion}

The consequence of our findings leads to questions about the necessity of machine learning approaches for some tasks. It is well known that if a machine learning model is optimized for one region or modality, it is not optimal for other modalities and body regions. In a recent study, this issue was raised for the registration task \cite{jena2025deep}, and it was shown that traditional methods have a better generalization ability when no additional supervision is used.

Consistent Point Matching has more potential than what is presented here, such as getting organ label for a query point. The algorithm could be extended to a full registration algorithm using multiple points. However, brute-force scaling does not yield a practical algorithm. Additionally, the current search operation considers only displacement values. In some body regions, such as in extremities, orientation is also important for finding the best matches. Thus, incorporating six degrees of freedom (three translational and three rotational parameters) in the search operation would increase the chance of finding the correct corresponding point. Further studies could investigate these more challenging anatomical regions.

%Consistent Point Matching has more potential than what is presented here, such as getting organ label for query point. The algorithm could be extended to a full registration algorithm using multiple points. However, brute-force scaling does not yield a practical algorithm. Additionally, the current search operation considers only displacement values. In some body regions such as extremities, orientation is also important to find the best matches. Thus, incorporating six degrees of freedom in the search operation would increase the chance of finding the correct corresponding point. Further studies could investigate these more challenging anatomies. % Further studies could investigate more challenging anatomies, such as extremities. %Extending the algorithm to cover these possibilities is a topic for future work. 

\section*{Conclusion}

We have demonstrated the Consistent Point Matching method, which allows the identification of anatomically similar locations between pairs of volumetric medical images. Somewhat surprisingly, without any training or data requirements, Consistent Point Matching achieves state-of-the-art performance at high speed of computation and without the need for accelerator hardware, surpassing meticulously trained machine learning models while being generic across tasks, modalities and body parts. We demonstrated its effectiveness on four different datasets. Further applications may enable additional tasks on medical images.

\newpage

\bibliography{halidziya}

\end{document}